\documentclass[preprint,12pt]{elsarticle}
\usepackage[numbers]{natbib}
\usepackage{amssymb}
\usepackage{amsmath}
\usepackage{xcolor} 
\usepackage{tcolorbox} 
\usepackage{color, colortbl}
\usepackage{hyperref}
\usepackage{multirow}

\journal{Journal of Systems and Software}

\begin{document}
\begin{frontmatter}

\title{Trimming the Risk: Towards Reliable Continuous Training for Deep Learning Inspection Systems}

\author[1]{Altaf Allah Abbassi\corref{cor1}}
\ead{altaf-allah.abbassi@polymtl.ca}
\cortext[cor1]{Corresponding author}

\author[2]{Houssem Ben Braiek}
\ead{h.benbraiek@sycodal.ca}

\author[1]{Foutse Khomh}
\ead{foutse.khomh@polymtl.ca}

\author[2]{Thomas Reid}
\ead{t.reid@sycodal.ca}

\affiliation[1]{organization={Polytechnique Montreal},
                city={Montreal},
                country={Canada}}

\affiliation[2]{organization={Sycodal Inc.},
                city={Montreal},
                country={Canada}}

\begin{abstract}
The industry increasingly relies on deep learning (DL) technology for manufacturing inspections, which are challenging to automate with rule-based machine vision algorithms. DL-powered inspection systems derive defect patterns from labeled images, combining human-like agility with the consistency of a computerized system. However, finite labeled datasets often fail to encompass all natural variations necessitating Continuous Training (CT) to regularly adjust their models with recent data. Effective CT requires fresh labeled samples from the original distribution; otherwise, self-generated labels can lead to silent performance degradation. To mitigate this risk, we develop a robust CT-based maintenance approach that updates DL models using reliable data selections through a two-stage filtering process. The initial stage filters out low-confidence predictions, as the model inherently discredits them. The second stage uses variational auto-encoders and histograms to generate image embeddings that capture latent and pixel characteristics, then rejects the inputs of substantially shifted embeddings as drifted data with erroneous overconfidence. Then, a fine-tuning of the original DL model is executed on the filtered inputs while validating on a mixture of recent production and original datasets. This strategy mitigates catastrophic forgetting and ensures the model adapts effectively to new operational conditions. Evaluations on industrial inspection systems for popsicle stick prints and glass bottles using critical real-world datasets showed less than 9\% of erroneous self-labeled data are retained after filtering and used for fine-tuning, improving model performance on production data by up to 14\% without compromising its results on original validation data.
\end{abstract}

\end{frontmatter}

\section{Introduction}
\label{sec:introduction}
Traditionally, visual inspections of manufactured products were conducted by human operators, resulting in a process that was error-prone, time-consuming, and resource-intensive \cite{app12042250}. To mitigate these challenges, rule-based machine vision inspections were introduced. These systems employ computer vision algorithms, such as blob or edge detection \cite{Lupi2023}, to identify defects by comparing actual products to ideal patterns developed by domain experts. While these methods automate the inspection process, they transfer the burden from operators to domain and vision experts. The emergence of new defects, changes in the manufacturing process, or deviations in the products necessitates substantial reprogramming and expert involvement to update algorithms or predefined patterns \cite{MENDE2022305}, highlighting the need for more adaptive and efficient inspection solutions.

To reduce the reliance on hand-crafted designs, industries have progressively adopted machine learning (ML) and deep learning (DL) algorithms for automated visual inspection systems \cite{Elhoseny2022, semeniuta2018towards}. This shift towards data-driven approaches offers a more adaptive and efficient method for identifying manufacturing product defects \cite{s19183987, Zheng2021}.  However, these statistically learned models may struggle with production inputs that differ significantly from the in-distribution (ID) data used during training, causing silent performance degradation after deployment \cite{hashmani2019accuracy, rabanser2019failing}. Production manufacturing facilities are often dynamic and consequently prone to many environmental variations, such as unexpected changes in luminosity, which add further complexity \cite{kahraman2022dynamic}. Creating a comprehensive training dataset that covers all potential scenarios is not feasible, due to the inherent uncertainty of such variations and their uniqueness \cite{10.1002/widm.1327}. Hence, automated maintenance of DL-powered inspection systems often involves Continuous Training (CT), which regularly adjusts models after deployment based on recent production data \cite{MLSYS2022_069a0027}. To be effective, CT demands fresh training samples that follow the original data distribution and are accompanied by ground truth labels. Using a model's predictions as self-generated labels can result in incorrect labels being incorporated, as even well-trained models may not perform perfectly when faced with novel inputs in real-world settings \cite{chen2022debiased}.
The goal is to minimize these failures to achieve high validation scores and release a viable model for the given problem. Relying on self-generated labels for difficult corner cases in production is likely to introduce errors. Additionally, production data can be subject to distribution drift, where novel inputs deviate from the ID. Drifted inputs may result in erroneous model behaviors, such as false overconfidence \cite{bayram2022concept}, and CT on drifted data—even with ground truth labels—does not guarantee that a retrained model will be viable. This is because all ML steps and hyperparameters were engineered and validated based on the ID data. Therefore, CT should be conducted on novel datasets that are correctly labeled and belong to the ID distribution or closely similar distributions \cite{kim2022theoretical}. Failure to meet these CT prerequisites may result in model performance degradation over time \cite{10.1259/bjr.20220878}.
In response to these challenges, we develop a robust CT-based maintenance approach that updates DL models using reliable data selections through a two-stage filtering process.
This aims to mitigate the risk of incorrectly labeled or drifted inputs being included in the dataset used for CT. Specifically, the confidence filter rejects all inputs with weak prediction confidence scores, as the model inherently discredits its self-generated labels for them. At this initial filtering stage, our objective is to eliminate corner cases that may exist on the boundaries of the ID and could still pose challenges for the model to generalize. 
The selected inputs with reliable confidence scores then pass through a follow-up drift filter that aims to filter out inputs that are substantially shifted from the ID data. Due to the complexity and high dimensionality of imaging data, capturing these adverse shifts involves a preliminary step of dimension reduction. To achieve this effectively, we combine two complementary embeddings to derive comprehensive and representative profiles for the images: (i) a Variational Auto-Encoder (VAE) to extract the most relevant semantic features from the images (i.e., retains what characterizes a given image from the other images within the data distribution), and (ii) a pixel histogram to generate a pixel profile as a numerical vector representing the image pixel-value characteristics, capturing tonal, color, and shade information. Combining these semantic and pixel-value profiles into an image embedding creates a new low-dimensional space. This space allows for computing distances between embeddings and efficiently learning statistical models. Typically, the ID data is available during model engineering, and data drift tends to occur after deployment in the production environment \cite{polyzotis2019data}. To simulate potential drifts, random data transformations with varying severity levels are commonly used. However, there is no guarantee that these transformed data are representative of all possible data drifts \cite{vernekar2019out}. Therefore, we train a one-class classifier on the ID image embeddings to capture their patterns and distinguish them from out-of-distribution (OOD) image embeddings. The transformed data are used to validate that the optimized one-class classifier can indeed reject substantially shifted inputs based on their embeddings.

Subsequently, our proposed CT-based maintenance approach proceeds with a fine-tuning of the DL model on the filtered inputs while validating on a combination of recent production and original training datasets. This strategy mitigates catastrophic forgetting, i.e., degradation of performance on the original distribution data, and ensures the model adapts effectively to new operational conditions.

Our proposed two-stage filtering approach minimizes the risk of retraining the model on non-reliable data that are either incorrectly labeled or substantially shifted from the original training data distribution. We evaluate our approach on two real-world industrial use cases: the first examines potential defects in Popsicle stick prints, and the second investigates defects that may occur during the production of glass bottles. An assessment on critical real-world scenarios involving data drifts shows that our CT-based maintenance retains only 8\% of incorrectly auto-labeled instances for fine-tuning the original model, and the updated model maintains its performance on production data but also enhances it by up to 14\%, without compromising its initial results on the original validation datasets.
The rest of this paper is structured as follows: Section \ref{sec:background} provides a comprehensive background of relevant concepts to our approach, while section \ref{sec:methodology} delves into the details of our two-stage filtering approach for production data and CT workflow. Evaluation results are presented in Section \ref{sec:evaluation}. Section \ref{sec:related_work} discusses the relevant related works, and Section \ref{sec:threats_to_validity} presents potential threats to validity.

\section{Background}
\label{sec:background}
This section provides background information about data drift, automated labeling, and continual learning.\\
\textbf{Data Drift} Production data distribution may be subject to changes over time, which can be gradual, sudden or seasonal \cite{GUO20221}, affecting the statistical properties or characteristics of the input data \cite{ackerman2021automatically}. The dynamic nature of production environment causes data to shift from the original data distribution \cite{mai2021online}. The shift may reveal the occurrence of new defects (uncovered by the training data) \cite{zhang2023toward} or the rise of a new domain \cite{tomani2021post}, which includes new background, blur, noise, illumination, occlusion \cite{lin2019concept, sun2021convolutional}. Data drift adversely influences the model performance but not discrediting its learned patterns and inductive bias \cite{ijerph19063641}. This is contrary to concept drift, which describes data evolution that invalidates the current model \cite{rahmani2023assessing}. \\
\textbf{Automated Labeling}\label{subsub:datalabeling}
Manual data labeling is often criticized as laborious and time-consuming step in ML engineering pipeline, and it can also become costly particularly if it necessitates subject-matter experts \cite{zhang2022survey}. To cut costs, ML models can be used alongside humans in a semi-supervised approach \cite{pmlr-v133-desmond21a}, where the model proposes annotation suggestions to humans aiming to speed up the labeling process. In a similar fashion, Active Learning (AL) is employed to optimize labeling efforts by selecting only the most informative datapoints for manual labeling \cite{ren2021survey}. Despite reducing labor costs and time, these human-in-the-loop approaches cannot be fully automated. Alternatively, weak supervision approaches such as Snorkel strategy \cite{ratner2017snorkel} are proposed, which use programmatic labeling rules to auto-label most of the data based on heuristics, patterns or domain knowledge \cite{Ratner2020}. In addition to being not trivial and complex to create, these labeling functions can only cover a subset of data \cite{bach2019snorkel}, so humans must be partially in the loop for iterative improvement and complementing the labeling functions. 
Another straightforward option is to use an ML model trained on labeled data for labeling unknown data \cite{zhang2021survey}. This approach is susceptible for corrupted labels caused by model errors. To reduce these errors, some strategies rely on model confidence score \cite{xiong2023confidence}. \\
\textbf{Continual Learning (CL)} In the context of ML, CL is a general concept that enables continuous and adaptive learning of production environment by enabling autonomous incremental acquisition of knowledge \cite{parisi2019continual}. CL techniques such as lifelong learning and incremental learning aim to adapt the model in a context of non-i.i.d (independent and identically distributed) data streams \cite{banerjee2023streaming}. Specifically, CT, or conventional CL, proposes to retrain the model on a regular basis \cite{9792270} in an offline manner on the newly collected data \cite{s22239298}. As a result, CT-enabled systems can accumulate knowledge that will improve their ability to handle potential future shifts in data \cite{suarez2023survey} while preserving the knowledge gained from the initial dataset. There are different approaches implementing the CT that can be categorized based on trigger: periodic retraining, performance-based retraining, or data-driven retraining and on demand \cite{9792270}. When it comes to the used data, two main categories emerge: training the model from scratch on all data \cite{Prapas2021} or transfer learning, in which an old model is fine-tuned to adapt to new data \cite{iman2022expanse}.

\section{Proposed Approach}
\label{sec:methodology}
In this section, we describe the different steps required to co-design our proposed two-stage filtering approach for reliable self-improving visual inspection system. 
\subsection{Characterization of Risky Data}
Supervised ML models are commonly estimated via empirical risk minimization (ERM)~\cite{goodfellow2016deep}, a principle that considers minimizing the average loss on observed samples of data, as an empirical estimate of the true risk, i.e., the expected true loss for the entire input distribution. The average cost reduction leads to greedily absorbing the patterns that hold on the majority of training instances, resulting in a biased model that is prone to outliers and unfair to minorities \cite{li2021tilted}. Hence, the first category of risky data is the underrepresented inputs that are most likely to be erroneously classified by the model due to their infrequent occurrences in the ID data. Furthermore, ERM assumes that training and test data are identically and independently distributed (a.k.a. i.i.d. assumption) \cite{yuan2022towards}, which explains the use of held-out validation data as a proxy for unseen data points. However, data drifts often occur in production environments for many reasons, such as naturally-occurring shifts (unfamiliar background, motion blur, random noise, unexpected illumination, occlusion, etc.) \cite{lin2019concept, sun2021convolutional, nikolov2021seasons}, new emerging defects, old defect pattern variations, or deviations in defect statistics \cite{zhang2023toward}. As the model may not generalize on drifted inputs, its overall predictive performance could suffer \cite{zhang2023map, MENDE2022305}. Thus, the second category of risky data consists of OOD inputs, i.e., data that belong to the true distribution (i.e., operational domain of the inspection system), but they are absent from the training and validation datasets due to selection bias. Finally, the third category of risky inputs consists of anomalous data that may incidentally arise in real-world production situations, but they do not fall under the inspection system's foreseeable operating conditions \cite{kammerer2019anomaly}, and therefore, should be identified and excluded.   
\subsection{Confidence-based filtering stage}
\label{subsec:conf_filter}
In ML, confidence scores represent the model's estimated probability that a given input belongs to a particular data class. In other words, these scores represent how confident the model is in its predictions \cite{le2023explaining}. Higher confidence scores indicate a higher level of certainty in determining the class of a given input, whereas lower scores suggest uncertainty \cite{wu2023sentistream}. However, the inherent complexity of neural networks, coupled with the main objective of ERM being to maximize classification accuracy during training, contributes to the lack of calibration in model confidence outputs \cite{guo2017calibration}. The calibration aligns the predicted confidence scores with the actual correctness rates, i.e., a model that is perfectly calibrated will predict a confidence score of $0.8$ for samples with a correctness rate of $80\%$ \cite{desai2020calibration}. In particular, calibrating confidence scores is crucial to gauging the reliability of predictions and quantifying the model's uncertainty \cite{guillory2021predicting}. Confidence calibration techniques can be used at different stages: training, post-training, and inference \cite{silva2023classifier}. Regularization, during-training calibration, is insufficient dealing with overconfidence \cite{guo2017calibration}. For ensemble approach, during-inference calibration are computationally expensive and time-consuming, as it involves multiple models \cite{ye2023efficient}. Post-training methods have been investigated with temperature scaling proving to be particularly simple and effective \cite{guo2017calibration}. Indeed, a temperature parameter, $T > 0$ is used to re-scale logits before applying softmax, $\text{softmax} = \frac{\exp(z / T)}{\sum_i \exp(z_i / T)}$. The $T$ is adjusted on the validation logits returned by the trained model by minimizing the Expected Calibration Error, which quantifies difference between prediction probabilities and real probabilities. In our approach, we used temperature scaling, which we found to be reliable and less prone to errors. It is also aligned with our continuous improvement philosophy, as its effectiveness can be enhanced with more validation data that helps in accurately estimating the temperature parameter and further improving the model's confidence scores. Overconfident situations, which are common in modern neural networks \cite{wei2022mitigating}, where probabilities frequently skew to $0$ or $1$, can be mitigated by $T > 1$. Nonetheless, even calibrated confidence scores should be interpreted with caution, as they may fail to capture epistemic uncertainty in scenarios like data drift \cite{ovadia2019can} and may suffer from reliability degradation. Therefore, we develop our first stage of data filtering based on confidence scores that targets the rejection of the underrepresented inputs. As this particular category of risky data belongs to ID but with less frequency, we believe that their corresponding calibrated scores would be relatively low due to the model's low accuracy on their input space regions \cite{feutry2019simple}. Our confidence-based filter requires setting a single parameter—a threshold—to segregate low-confidence predictions from unseen inputs and prevent their inclusion in the self-labeled data. Tuning this threshold is crucial and involves a tradeoff: a higher threshold reduces false positives (incorrectly classified inputs), but increases false negatives (correctly classified inputs mistakenly rejected). Alternatively, a lower threshold maximizes true positives (correctly classified inputs), lowering false negatives but increasing false positives. Despite seeming counterintuitive, a lower threshold can improve overall results by allowing more inputs to proceed to our subsequent filtering stage, which can effectively remove the residual risky data while preserving enough correct samples for continuous training. Therefore, we propose to address the acceptable performance level of the inspection system as a primary consideration. Then, we select the lowest threshold where the achieved performance meets this criterion. This procedure enables us to tolerate false predictions in order to mitigate the risk of rejecting correct predictions, all while ensuring that we do not fall below our target performance level.
\subsection{Distribution-based filtering stage}
A major drawback of ML models, which is further accentuated by the rise of deep neural networks with superior learning capacity, lies in their propensity to always generate an output, as long as the inputs are provided in proper numerical form, even if such input is meaningless in the domain where the DNN is supposed to operate \cite{rabanser2019failing}. Likewise, we observe the inability of the models to distinguish the ID inputs from the ODD ones. This has led to the emergence of out of distribution/drift detection approaches that aim to complement the predictive models and overcome their limitations regarding self-recognition of their valid input domains. Following are the details of our proposed distribution-based filter that reveals OOD images in the context of visual inspection, i.e., the inputs do not fall within its operational domain.
\subsubsection{Rich image embeddings}
In the context of visual inspection system, high-resolution images inherently present a challenge due to their high-dimensional nature, making it difficult to measure similarity and dissimilarity between them \cite{aggarwal2001surprising}. As a solution, semantic feature extraction has gained traction in practice, which involves employing diverse methods, ranging from statistical techniques like Principal Component Analysis (PCA) to DL-based approaches like autoencoders \cite{yu2020spatial}. The extraction of meaningful and condensed representations for the input images streamlines the data analysis by capturing essential information while reducing the dimensionality \cite{khalid2014survey}. In our approach, we leverage VAE\cite{kingma2013auto} to derive image embeddings that retain semantic variations across instances. As deep generative models, VAEs comprise two fundamental components: the encoder and the decoder. The encoder takes an input image and maps it to a latent representation distribution, then used by the decoder to reproduce the same image as output \cite{kingma2013auto}. The learned latent representation encapsulates a valuable and condensed embedding for semantic image profiling because it encodes the images into a smooth, continuous, lower-dimensional space while preserving essential information to differentiate between them based on their semantic content \cite{soin2022chexstray}. 
Contrary to conventional autoencoders, VAEs employ stochastic encoding representation that captures the underlying distribution of the latent space \cite{ye2021learning}, preserving the diverse aspects of the encoded images. Although this probabilistic aspect improves robustness, VAE models do have limitations. These models may suffer from posterior collapse, which occurs when the VAE encoder fails to generate accurate input embedding \cite{lucas2019don}. This leads to embedding overlap between ID and OOD inputs, and in more severe cases, the latent space can become independent of the input data.
To address these limitations and enhance encoding capacity, we opted for fine-tuning VAE rather than training from scratch. This approach facilitates the learning of robust, high-level embedding representations. To create this pre-trained VAE, we opt for MVTec AD \cite{Bergmann2021}, which is a dataset for benchmarking industrial inspection models. It includes 5354 high-resolution images across fifteen object and texture categories, with each category containing defective and defect-free images. By training the VAE on this rich defect dataset first, high-level domain patterns can be acquired, which are then transferred to the use-case-specific VAE, reducing the risk of posterior collapse and improving its OOD representation. 
As we aim as well, to distinguish the shifted data from ID data, VAE-generated image embeddings should be sensitive to semantic shifts. To accomplish this, we systematically simulate both moderate and significant shifts by applying diverse image transformations such as blurring, sharpening, and brightening at varying intensity levels (more details in Section \ref{subsec:occ}). Subsequently, we measure the disparity between the embeddings of original images ("original embedding") and the embeddings of synthetically-shifted images ("shifted embedding") in order to assess their discriminative power. In fact, we calculate the average of the pairwise distances between all the original and shifted embeddings. We use the $L_{\infty }$ as a distance measure that computes the maximum absolute difference, while averaging offers an overall assessment of the discrepancies between the two data distributions. As disparity levels increase, the derived image embeddings become more accurate at separating ID instances from their synthetically-shifted counterparts. While VAE embeddings are effective in capturing semantic differences among images, they may overlook fine-grained pixel value distributions \cite{seyfioglu2024diffuse}, as it can manifest as random noise, color changes, and semantically-preserving variations. Nonetheless, such pixel-level variations can negatively affect the neural network behavior and leads to misclassification, as evidenced by the studies on adversarial attacks \cite{akhtar2018threat}. To consider pixel-value variations in our image embeddings, we developed an histogram profiler to capture the nuances of pixel value distributions. We start by calculating a condensed histogram that aggregates pixel occurrences within different buckets (ranges of values) for each color channel. Subsequently, the occurrences are normalized by dividing them by the total number of pixels (i.e., height $\times$ width), then, concatenated them sequentially in the order of the red, green, and blue channels. Therefore, our compact rich embedding is the concatenation of the VAE encoding profile and the histogram-based image profile. 
\subsubsection{Classification-based OOD detection}\label{subsec:occ}
Once we construct the compact embedding space, the next step is to learn the patterns and characteristics of the ID embeddings compared to OOD embeddings. The use of supervised learning algorithm poses a significant challenge, arising from the necessity of explicitly identifying unknown potential future forms of distribution drift \cite{hanqing2020}. As an alternative, statistical testing can assess whether production data embeddings belong to the same distribution of the training data embeddings\cite{alberge2019detecting} \cite{ackerman2021automatically}. Our derived embeddings serve as condensed representations of images, yet they may consist of hundreds of elements that do not conform to any specific underlying distribution. Thus, multivariate statistical tests, known for their sensitivity to slight changes \cite{polyzotis2019data}, are rendered ineffective in our use case \cite{lavergne2008breaking}. 
Another avenue within the realm of unsupervised anomaly detection involves the utilization of one-class classifiers (OCC), such as Isolation Forest (IF) or One-class Support Vector Machine (OC-SVM) \cite{farzad2020unsupervised}. These ML algorithms are trained on samples belonging to a single class. The resulting model is capable of delineating a boundary decision surrounding the training data distribution. At the inference, the model predicts whether a new input belongs to the learned data distribution or not \cite{khan2014one}. These OCC models handle multi-dimensional data well, and they do not require OOD training samples. Typically, anomalous or drifted inputs are reserved for testing to evaluate the model performance. These one-class classifiers have shown their effectiveness at detecting anomalies and OOD instances \cite{farzad2020unsupervised, 9580016}. Our research focuses on identifying risky data to filter them out because the model may produce false (random) predictions, and even with correct predictions, their inclusion into training samples may adversely alter the model's inductive bias \cite{kim2022theoretical}. These risky data can manifest as unexpected or OOD inputs, as well as instances that are under-represented in the training data distribution. In conventional DL model engineering scenarios, all available data is typically considered during the design process. Consequently, it becomes challenging to anticipate drifted data, often requiring their collection post-deployment. To approximate shifted production data, we employ synthetic data transformations outlined below to generate shifted versions of the original samples. These transformations offer flexibility in introducing varying degrees of shift.\\ 
\textbf{Blurring} is achieved through the application of a Gaussian filter. The degree of blurring, reflecting the transformation's intensity, is adjusted by tuning the standard deviation ($\sigma$) within the Gaussian filter, thereby managing the spread of the underlying Gaussian function. A higher $\sigma$ indicates a more aggressive transformation. A $\sigma$ value of  $0$ renders the original image, and there is no upper limit for this value.\\
\textbf{Brightening} is achieved through the utilization of a Brightness adjustment filter, employing the brightness factor, denoted as $\beta$, as an intensity control. This ensures uniform brightness enhancements over the entire image. The higher $\beta$ the more pronounced and aggressive the transformation becomes. Notably, $\beta$ value of 0 results in a black image, a $\beta$ of 1 renders the original image, and there is no upper limit for this value.\\
\textbf{Sharpening} is accomplished through the application of an Unsharp Masking transformation. This involves creating a degraded version of the image and then subtracting it from the original to obtain a sharpened image. The degree of sharpness is controlled by a sharpening factor, denoted as $s$ , where a higher value corresponds to a more aggressive transformation. A $s$ value of $0$ produces a blurred image, while a factor of $1$ maintains the original image. \\ 
Identifying the maximum level of distortion tolerated by a given model poses a challenge, as transformed inputs within a certain intensity range may either drift or remain acceptable based on the original data. To simplify the task, we opt for categorizing the synthetic data transformations into two distinct groups: moderate and harmful data shifts, relying on their intensities. Indeed, moderate shifts closely resemble the conventional data augmentation techniques~\cite{shorten2019survey} employed to enrich training datasets with inputs derived within the in-distribution, thereby increasing model generalizability. Conversely, harmful shifts produce synthetic OOD data that are beyond the model's limits, but may result from unfavorable production conditions. They are prone to inducing erroneous predictions and instabilities during retraining, ultimately resulting in deviations from the model's intended behavior. To determine the intensity level of each shift, we initiate by employing default intensities commonly used in data augmentations~\cite{shorten2019survey}, then refining them through qualitative validation (i.e., human-in-the-loop). Indeed, we manually examine the input semantics before and after each transformation w.r.t the data distribution in order to ensure the preservation of relevant information necessary for the prediction task. This human-in-the-loop validation process draws upon existing research on semantically-preserving transformations~\cite{deephunter, deepevolution}. To construct the distribution-based filter, we train an IF model on our compact embeddings of original training samples. Then, we assess the trained IF model on the produced synthetic dataset. The assessment reposes on estimating the acceptance and rejection rates for both categories: moderately-shifted and harmfully-shifted inputs. The optimal IF model is one that can simultaneously maximize the rejection rate for harmful shifts and the acceptance rate for moderate shifts. Hence, our distribution-based filter can effectively reject all drifted data (represented by harmful shifts), wherein the inspection system is prone to erroneous predictions, while still accepting incoming inputs closely resembling the in-distribution data (represented by moderate shifts). 
\subsection{Continuous Training-driven Maintenance}
In this present research, we implement a reliable CT-driven maintenance approach for DL-based inspection systems. For such systems, the vanilla CT with self-generated labels can be effective when production data are aligned with the training samples' distribution \cite{kim2022theoretical}. However, real-world industrial environments introduce dynamic variations such as changes in luminosity, vibration patterns affecting object positions, and machinery depreciation-induced behavioral shifts \cite{kahraman2022dynamic}. In response to these environmental variations, we expect DL-based inspection systems to maintain a certain performance level due to their inherent robustness to data distortions and noise that do not significantly compromise semantic information. However, vanilla CT with self-generated labels fails to enhance and even maintain inspection model performance amidst such environmental fluctuations, as the number of false predictions increases and incoming inputs may substantially deviate from the in-distribution samples \cite{kim2022theoretical}. Therefore, a preliminary filtering step is crucial to exclude risky data points (i.e., false predictions or drifted inputs) from the auto-labeled datasets supplied to CT. Initiating CT can be prompted by meeting a specified condition, such as after a pre-defined time period, performance degradation triggers, on-demand retraining, or a combination of these conditions \cite{9792270}. In our context of supervised industrial inspection, the monitoring of performance degradation depends on groundtruth labels and the on-demand retraining requires ML expert oversight. Hence, periodic CT can be streamlined by synchronizing the intervals with production pauses and scheduling maintenance within a time window that allows for the collection of an adequate number of production entries. In between CT cycles, the DL-based inspection system logs all inputs alongside their model-generated predictions, i.e., labels and calibrated confidence scores. Upon CT triggering, a confidence-based filter is employed to discard inputs associated with low (calibrated) confidence scores, specifically those below a predefined threshold carefully tuned on testing datasets. Subsequently, the remaining inputs with adequate confidence levels undergo a distribution-based filter. This filter generates rich embeddings of the inputs, which are then fed into an Isolation Forest (IF) model optimized to reject any embedding substantially shifted from the in-distribution embeddings. As a result, the final retained inputs are more likely to be correctly classified and sufficiently similar to the original samples (i.e, no significant deviation from the in-distribution). Thanks to transfer learning, the CT-driven maintenance of DL-based inspection system does not require the retraining of the model from scratch on the entire dataset, i.e., combining the original and production entries. Instead, it operates directly on the last production model, considering it as the base (pretrained) model, and fine-tunes it exclusively on the filtered production data using the same hyperparameters, except for decreasing the maximum number of epochs. This reduction helps mitigating the risk of catastrophic forgetting phenomenon, where the model overfits to the new samples to the extent of losing its original performance. Furthermore, a merged validation dataset that includes both of original and novel validation data (gathered after deployment) also contributes to prevent catastrophic forgetting. Indeed, the resulting comprehensive assessment ensures that the model adequately handles both new and past entries, contributing to sustained performance across CT cycles. It is worth noting that the new validation dataset was derived by partitioning the production data according to the same training/validation split ratios utilized in the initial training process. This approach ensures a proportional representation of the current and historical sources of training data for continuous training (CT).

\section{Evaluation}
\label{sec:evaluation}
In this section, we detail our evaluation setup including use cases, models, metrics, and procedures. Next, we present the evaluation results through examining four research questions. 
\subsection{Experimental Setup}
\subsubsection{Use Cases}
\begin{figure}[]
  \centering
  \begin{minipage}[b]{0.24\textwidth}
    \centering
    \includegraphics[scale=0.36]{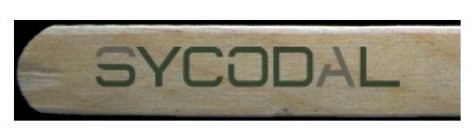}
  \end{minipage}\hfill
  \begin{minipage}[b]{0.24\textwidth}
    \centering
    \includegraphics[scale=0.36]{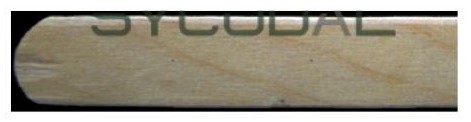}
  \end{minipage}
  \caption{Good vs Decentralized Popsicle Stick Print}
  \label{fig:pop_illus}
\end{figure}

\begin{figure}[]
  \centering
  \begin{minipage}[b]{0.2\textwidth}
    \centering
    \includegraphics[scale=0.08]{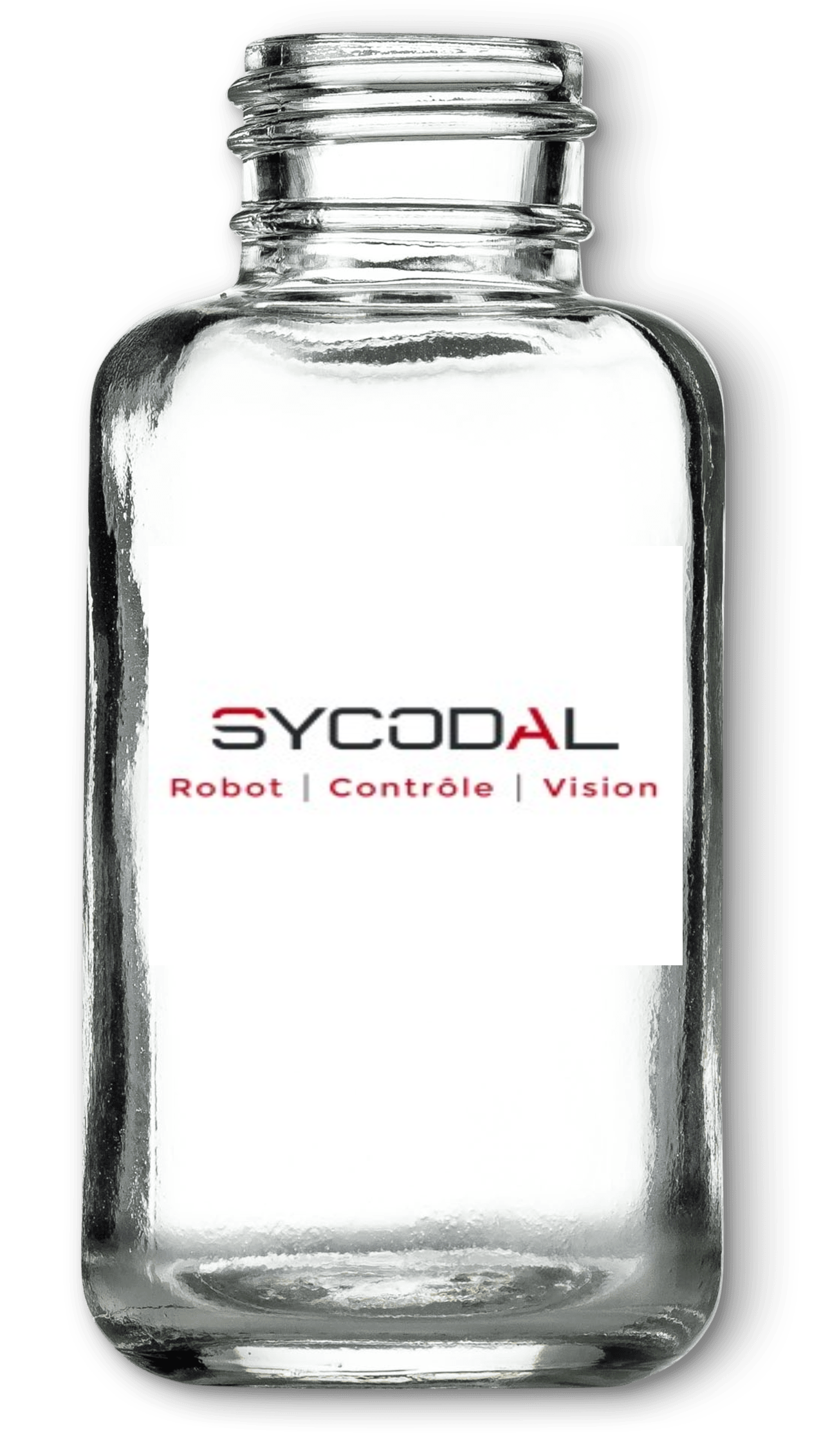}
  \end{minipage}\hfill
  \begin{minipage}[b]{0.2\textwidth}
    \centering
    \includegraphics[scale=0.08]{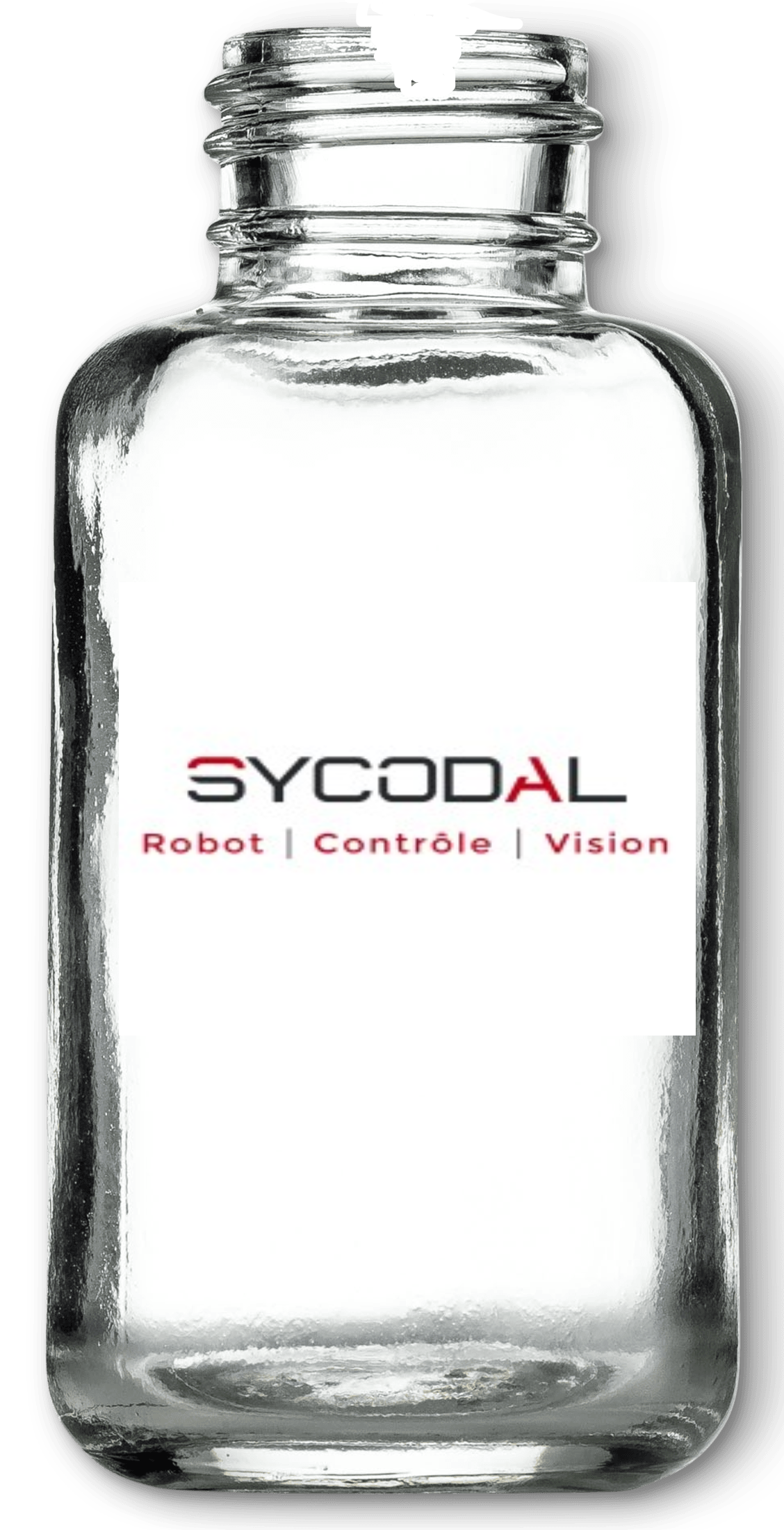}
  \end{minipage}
  \caption{Defective vs Non-Defective Glass Bottle}
  \label{fig:bottle_comparison}
\end{figure}

The datasets for our use cases consist of proprietary images from industrial partners, which we cannot share due to non-disclosure agreements. Instead, we describe the use cases, datasets, and data collection and preparation strategies. \\
\textbf{Popsicle Stick Prints (POP).} The task involves the quality inspection of logos and text prints on Popsicle wood sticks based on a single top camera view. The inspection focus on the precision and clarity in the logo and the overall readability of text, e.g., a pale-colored logos or text prints should be considered as a defect in printing. An illustration of Popsicle stick is presented in Figure \ref{fig:pop_illus}.
\\
\textit{POP Original Dataset.} It includes 1212 images, of which 700 represent good stick prints, and 512 exhibited defective stick prints. These images, captured at a resolution of 500x115 pixels, were obtained during the development phase and were divided into 80\% training and 20\% validation samples.\\
\textit{POP Production Dataset.} It represents a dataset collected in production environment (on-site testing phase) within a critical window time, during which the model experiences a severe degradation in its predictive performance. This critical production dataset contains 356 images.\\
\textbf{Glass Bottles(GB).} This task is concerned with the quality inspection in glass bottles filled in packaging boxes based on a single top camera view. Depending on the packaging strategy, bottles can be filled in two orientations: either upright, revealing the bottle neck, or inverted, showcasing the bottom of the glass bottle. Whenever one bottle is the inverse of what is expected, it is regarded as a defect. For bottle defects, we found the occurrence of broken bottle finish when the bottle exposes its neck on the top, and the occurrence of cracked bottom when the bottle is turned upside down. An illustration of these defects is presented in \ref{fig:bottle_comparison}.
\textit{GB Original Dataset.} It comprises 1195 images, wherein 597 depicted good bottles without any defects, while 598 showcased bottles with defects either at the bottom or the finish of the neck. The resolution of the images is on average of 224*224 pixels. Indeed, these images of good/defected bottles are collected by cropping an original image of 6-glass bottle packaging box, captured at a resolution of 1024*1024, into 6 sub-images containing each a bottle.\\
\textit{GB Production Dataset.} Similar to the POP use case, we collected at on-site testing phase, a subset of critical dataset on which the model poorly behaves. We obtained a total of 455 images gathered within a specific window time.
\subsubsection{Image Transformation Settings} \label{subsub: data_generattion process}
We set the following factors after validating by experts for the image transformations described in \ref{subsec:occ}:\\
\textit{Moderate Shift.} $\beta$: (1.5 and 1.8), $s$: (5 and 20) for both use cases, and for $\sigma$ :(2 and 3) and (0.8 and 1) for POP and GB use case respectively.  \\
\textit{Harmful Shift.} $\beta$: (2.5 and 3), $s$: (100 and 500) and $\sigma$: (5 and 10) and  for both use cases.\\
We apply these transformation to generate two datasets: moderate and harmful synthetic samples for VAE pre-training and OCC validation. \\
\subsubsection{Models} 
\textbf{Binary Classification.} For both use cases, the inspection system relies on a binary classifier leveraging ResNet \cite{7780459} pretrained model sizes: $18$, $50$ and $101$. Best results are given by \href{https://huggingface.co/timm/resnet18.a1_in1k}{ResNet18}. The optimal hyperparameters are the following: learning rate of $10^{-4}$ and a batch size of $32$, spanning $200$ epochs with early-stopping, Adam as optimization algorithm and Cross Entropy as loss function. For fine-tuning during CT, we reduce the epochs to $50$ to avoid catastrophic forgetting as only the entries from the last time window are used.\\
\textbf{VAE.} It comprises an encoder with three convolutional layers succeeded by max-pooling layers, followed by a flatten and two hidden layers and a decoder mirroring the reverse structure of the encoder. All layers employ the Rectified Linear Unit (ReLU) as neuron activation. We use Adam as optimizer and Mean Squared Error as loss function, as well as a learning rate of $10^{-4}$, a batch size of $32$, and spanning $200$ epochs. We explore various latent space dimensions, namely $64$, $128$, $256$, $512$, and $1024$.\\
\textbf{OCC.} We select Isolation Forest (IF) with tuning hyperparameters using: (i) proportion samples used range from $0.1$ to $0.9$ with a step of $0.1$; (ii) contamination proportion range from $0.01$ to $0.4$ with a step of $0.05$, maximum number of features range from $0.1$ to $1.0$ with a step of $0.1$. 
\subsubsection{Evaluation Procedure}
We conduct multiple experiments, repeated at least 5 times to secure robust set of results. Below are the the metrics and the procedure used.\\
\textbf{F1-score (F1)} is the harmonic mean of precision and recall to balance between these two metrics.\\
\textbf{\%True Acceptance (TA)} denotes the ratio of correctly-classified instances that successfully pass a given filter.\\
\textbf{\%False Acceptance (FA)} represents the ratio of incorrectly-classified instances that successfully pass a given filter.\\
\textbf{\%True Rejection (TR)} indicates the ratio of incorrectly-classified instances that are both rejected a given filter.\\
\textbf{\%False Rejection (FR)} indicates the ratio of correctly-classified instances that are rejected by a given filter.\\
\textbf{Dispersion.} consists of the average of the maximum absolute differences between two vector distribution, which can be formulated as follows: $\left \| U- V \right \| = avg(max_u (\sum_{v}^{} \left \| u - v \right \|_\infty ))$.\\
\textbf{Qualitative.} Data were sampled randomly from the various groups based on their label correctness and associated two-stage filter decisions: pass or reject. Then, we provide them to two domain experts for qualitative analyses of the possible causes of false predictions or rejections by filter. For both studied use cases, a summary of the analysis' outcomes is added to help understand the findings.\\
\subsection{Production Environment} The engineering team targets a performance of 95\% in terms of F1-score, but requires a minimum of 90\% of validation F1-score to deploy the model for testing in the production. However, following the deployment, the system experienced performance decrease, during which we collected production data. Table \ref{tab: classifier_per} summarizes model performance during the development process and on (critical) production data. 

\begin{table}[]
\centering
\caption{Binary Classifier Model Performance}
\begin{tabular}{|l|l|l|}
\hline
Use case & Validation F1 & Production F1 \\ \hline
POP     & 93.2\%       & 45.7\%       \\ \hline 
GB      & 93.3\%       & 76.6\%       \\ \hline 
\end{tabular}
\label{tab: classifier_per}
\end{table}
\subsection{Experimental Results} 
To evaluate our proposed approach, we studied the following research questions:
\subsubsection*{RQ.1: How effectively can confidence-based filtering be used to reject false predictions?} \hfill
\\
\textbf{Motivation.} Calibrated confidence scores can indicate how certain the model is in its predictions. We aim to assess their effectiveness in filtering out the inputs associated with unreliable model's outcomes.\\
\textbf{Method.} Given the established performance of 95\% F1-score, our procedure (Section \ref{subsec:conf_filter}) is to select the lowest threshold for confidence scores, where the model achieved the target score on the validation dataset. Then, we compute the true and false acceptance rate (TA and FA) as well as the true and false rejection rate (TR and FR) obtained by the designed filters on production data collected for both use cases. The assessment of the confidence-based filter effectiveness is conducted on the (critical) production dataset. This shows their capabilities as they were deployed in production when these critical inputs occurred.\\ 
\textbf{Results.} Table \ref{tab: conf_filter} shows, on one hand, the results of confidence-based filter on validation datasets after selecting the most adequate threshold. On the other hand, it reports the same metrics on production datasets that are degraded, i.e., the true acceptance rate dropped while the false acceptance increased by 5 times for POP and 10 times for GB. Nevertheless, the rejection rates are relatively low and stable among the experiments. This reinforces the overconfidence phenomenon where the model tends to assign high confidence scores even for unusual, novel data. A follow-up filtering stage is needed to further remove the passed inputs labeled by false predictions in order to prepare for a reliable CT.

\begin{tcolorbox} 
[colback=gray!8,colframe=gray!40!black]
\textbf{Finding}: Confidence-based filters tend to accept high proportion of inputs at production, despite the increased risk of false acceptance.
\end{tcolorbox}
\begin{table}[]
\centering
\caption{Confidence Filter Performance}
\begin{tabular}{|l|l|l|l|l|l|l|}
\hline
Use case              & Thresh.                & Dataset    & TA      & FA               & TR     & FR      \\ \hline
\multirow{2}{*}{POP} & \multirow{2}{*}{0.79} & Validation & 87.5\% & 3.3\%           & 2.1\% & 7.1\%  \\ \cline{3-7}  
                     &                           & Production & 65.7\% & \textbf{16.6\%} & 7.1\% & 10.6\% \\ \hline 
\multirow{2}{*}{GB}  & \multirow{2}{*}{0.58} & Validation & 91.9\% & 3.8\%           & 2.6\% & 1.7\%  \\ \cline{3-7} 
                     &                           & Production & 65.5\% & \textbf{30.5\%} & 2.2\% & 1.8\%  \\ \hline 
\end{tabular}
\label{tab: conf_filter}
\end{table}
\subsubsection*{RQ.2: Can the proposed embeddings serve as features to separate between shifted and original data inputs?}  \hfill 
\\
\textbf{Motivation.} The aim is to ensure that the designed embeddings can distinguish the incoming inputs based on their degree of deviation from the in-distribution data, i.e., original samples.\\
\textbf{Method.} First, we generate input samples form original dataset, moderately-shifted dataset, harmfully-shifted dataset. Then, we compute their respective embeddings using the proposed encoding techniques: VAE and Histogram. We vary the embedding space dimensions for VAE and Histogram by changing the latent space size and the pixel bucket size, respectively. Once the embeddings are computed, we estimate the dispersion between the original embeddings and the moderately-shifted embeddings, denoted O.-vs-M., and the dispersion between the original embeddings and the harmfully-shifted embeddings, denoted O.-vs-H., using the introduced measure. The higher the dispersion between the originals and shifted ones, the better the embedding method will be at capturing discriminative information.\\ 
\textbf{Results.} Table \ref{tab: vae_disp} reports the dispersion measurements for both VAE experiments: O.-vs-M. and O.-vs-H., with increasing latent space dimension, namely, 1024, 512, 256, 128, and 64. In similar manner, Table \ref{tab: hist_disp} shows the results of Histogram experiments with increasing bucket size, explicitly, 10, 20, and 30. As expected, both encoding techniques show a dispersion between the original and the moderately-shifted embeddings higher than the dispersion between the original and the harmfully-shifted embeddings. As a result, this demonstrates their ability to retain relevant information in order to capture potential distribution shifts. While all the proposed encodings can differ originals from the shifted versions, we found that the increasing of the space size, i.e., latent features or bucket size, causes the reduction of dispersion measures, which reflects a negative impact on the discriminative power of the created embeddings for both use cases. Therefore, we decided to proceed with the reduced dimensional size of both VAE and Histogram, 64 latent features and 10 bucket size, that showed the maximum dispersion between the two use cases. In addition, we will merge them into a concatenated feature vector to enrich the information for the one-class classification.

\begin{tcolorbox} 
[colback=gray!8,colframe=gray!40!black]
\textbf{Finding:} The proposed VAE and Histogram encoding methods produce dimensionally-reduced image embeddings, capturing distribution shifts effectively.
\end{tcolorbox}

\begin{table}[]
\centering
\caption{VAE Encoding Dispersion}
\begin{tabular}{|c|llll|}
\hline
\multirow{3}{*}{Latent Dimension} & \multicolumn{4}{c|}{Use case}                                                                                                            \\ \cline{2-5} 
                                  & \multicolumn{2}{c|}{POP}                                                & \multicolumn{2}{c|}{GB}                                       \\ \cline{2-5} 
                                  & \multicolumn{1}{c|}{O.-vs-M}            & \multicolumn{1}{c|}{O.-vs-H}            & \multicolumn{1}{c|}{O.-vs-M}             & \multicolumn{1}{c|}{O.-vs-H} \\ \hline
1024                              & \multicolumn{1}{l|}{1.45}          & \multicolumn{1}{l|}{1.73}          & \multicolumn{1}{l|}{2.43}           & 2.82                    \\ \hline
512                               & \multicolumn{1}{l|}{1.38}          & \multicolumn{1}{l|}{1.67}          & \multicolumn{1}{l|}{2.13}           & 2.43                    \\ \hline
256                               & \multicolumn{1}{l|}{1.44}          & \multicolumn{1}{l|}{1.85}          & \multicolumn{1}{l|}{2.74}           & 3.08                    \\ \hline
128                               & \multicolumn{1}{l|}{1.56}          & \multicolumn{1}{l|}{1.81}          & \multicolumn{1}{l|}{2.8}            & 2.98                    \\ \hline
64                                & \multicolumn{1}{l|}{\textbf{1.64}} & \multicolumn{1}{l|}{\textbf{1.95}} & \multicolumn{1}{l|}{\textbf{4.013}} & \textbf{4.49}           \\ \hline
\end{tabular}
\label{tab: vae_disp}
\end{table}

\begin{table}[]
\centering
\caption{Histogram Encoding Dispersion}
\begin{tabular}{|c|llll|}
\hline
\multirow{3}{*}{Bucket Size} & \multicolumn{4}{c|}{Use case}                                                                                                           \\ \cline{2-5} 
                          & \multicolumn{2}{c|}{POP}                                                & \multicolumn{2}{c|}{GB}                                      \\ \cline{2-5} 
                          & \multicolumn{1}{c|}{O.-vs-M}            & \multicolumn{1}{c|}{O.-vs-H}            & \multicolumn{1}{c|}{O.-vs-M}            & \multicolumn{1}{c|}{O.-vs-H} \\ \hline
10                        & \multicolumn{1}{l|}{\textbf{0.51}} & \multicolumn{1}{l|}{\textbf{0.82}} & \multicolumn{1}{l|}{\textbf{0.33}} & \textbf{0.56}           \\ \hline
20                        & \multicolumn{1}{l|}{0.36}          & \multicolumn{1}{l|}{0.55}          & \multicolumn{1}{l|}{0.26}          & 0.38                    \\ \hline
30                        & \multicolumn{1}{l|}{0.28}          & \multicolumn{1}{l|}{0.43}          & \multicolumn{1}{l|}{0.21}          & 0.29                    \\ \hline
\end{tabular}
\label{tab: hist_disp}
\end{table}

\subsubsection*{RQ.3: To which extent can the proposed distribution-based filter complement the confidence-based filter for more reliable input selection?} \hfill 
\\
\textbf{Motivation.} Even after calibration, confidence-based filters do not perform as well on validation datasets as they do on production datasets. By examining a follow-up filtering stage based on distribution, we aim to determine whether data drift is responsible for this reliability degradation.\\
\textbf{Method.} We follow the same method performed for the confidence-based filter evaluation. We assess the performance of the filter on the production dataset in terms of true and false acceptance rate (TA and FA), as well as true and false rejection rates (TR and FR). As both encoding techniques have shown promising results, we train the IF model on different image embeddings: VAE, Histogram, and VAE + Histogram (a vector concatenation), and we compare their respective performances as input features for the distribution-based filter. Training the IF model is carried out using the original samples but the validation is conducted on the transformed inputs, simulating moderate and harmful data shifts. After the assessment of distribution-based filter (alone) on the production dataset, we assess the performance of the end-to-end filter, sequentially applying the confidence-based filter first on the production dataset and then the distribution-based filter on solely the passed inputs. We can therefore study the complementarity of both filters.\\
\textbf{Results.} Table \ref{tab: drift_res} reports the performance results of distribution-based filter on the entire production data using different image embeddings for both use cases. As expected, the concatenation of the embeddings from the proposed two encoding techniques (VAE + Histogram) yields the best performance rates, as the one-class classifier has access to more discriminative and diverse information about the images to capture the in-distribution data characteristics. The VAE or Histogram encoding yields lower true and false acceptance rates when used alone for the POP use case. For the GB use case, they result in higher false acceptance rates and lower true acceptance rates. The concatenation of their embeddings, however, shows their complementarity and provides the highest true acceptance rates with relatively low false acceptance rates. If we compare the results of distribution-based filter in Table \ref{tab: drift_res} with the results of confidence-based filter in Table \ref{tab: conf_filter}, we found that the distribution-based filter tends to reject higher number of inputs than the confidence-based filter. As a result, both true and false acceptance rates are modestly reduced for POP use case, indicating that the data shifts are not pronounced in production data. In contrast, both true and false acceptance rates are substantially decreased for GB use case, suggesting that production data are indeed shifting. It is worth noting that despite the data shifts, true and false rejection rates are comparable, respectively, 25\% and 34\%, for GB use case. This demonstrates that the distribution shift does not induce systematically a false prediction, and the model tends to produce as many correct predictions as incorrect ones for inputs rejected based on their deviation from original data. Nevertheless, correctly-labeled, drifted data are considered unreliable for continuous training because they alter model patterns and cause catastrophic forgetting.

\begin{table}[]
\centering
\caption{Distribution-based Filter Results}
\begin{tabular}{|c|c|c|c|c|c|}
\hline
Use case              & Encoding         & TA                          & FA                           & TR                           & FR                           \\ \hline
\multirow{3}{*}{POP} & VAE Enc          & 18.6\%                      & 3.4\%                       & 20.3\%                      & 57.7\%                      \\ \cline{2-6} 
                     & (VAE + Hist) Enc         & 57.7\%                      & 13.4\%                      & 10.3\%                      & 18.6\%                      \\ \cline{2-6} 
                     & Hist Enc  & 25.4\%                      & 8.0\%                       & 15.7\%                      & 50.9\%                      \\ \hline
\multirow{3}{*}{GB}  & VAE Enc          & \multicolumn{1}{l|}{32.7\%} & \multicolumn{1}{l|}{14.1\%} & \multicolumn{1}{l|}{18.7\%} & \multicolumn{1}{l|}{34.5\%} \\ \cline{2-6} 
                     &  (VAE + Hist) Enc         & \multicolumn{1}{l|}{33.2\%} & \multicolumn{1}{l|}{7.7\%}  & \multicolumn{1}{l|}{25.1\%} & \multicolumn{1}{l|}{34.1\%} \\ \cline{2-6} 
                     & Hist Enc  & \multicolumn{1}{l|}{13.0\%} & \multicolumn{1}{l|}{16.5\%} & \multicolumn{1}{l|}{16.3\%} & \multicolumn{1}{l|}{54.3\%} \\ \hline
\end{tabular}
\label{tab: drift_res}
\end{table}


\begin{tcolorbox} 
[colback=gray!8,colframe=gray!40!black]
\textbf{Finding:} The VAE and Histogram methods can be combined to produce a rich, merged embedding that can serve as feature to learn a OCC effective in distinguishing between ID and OOD instances.
\end{tcolorbox}

Table \ref{tab: e2e_res} reports the performance results of the proposed two-stage filter, which combines both confidence-based and distribution-based filters, in order to pass only the inputs with confident predictions and likely belonging to the in-distribution. For both use cases, the true and false acceptance rates (TA, FA) are further reduced, resulting in (50\%, 8.85\%) and (31.42\%, 5.49\%) for, respectively, POP and GB. This means that the false labels, 8.85\% and 5.49\%, represent almost 15\% of the entire passed proportion data, 58.85\% and 36.91\% for, respectively, POP and GB. However, the two-stage filter manages to pass over half of the production data for the POP use case, but almost a third for the GB use case. The proportion of passed data is relevant because the data size has effect on the effectiveness of the subsequent model training iterations. In fact, the confidence-based filter passes more production data for GB than for POP use case, but the false acceptance rate was the half of the true acceptance rate, which shows a pronounced overconfidence issue (see Table \ref{tab: conf_filter}).The distribution-based filter alone leads to a low acceptance rate in the GB use case (see Table \ref{tab: drift_res}). The two-stage filter is affected directly by this reduction in acceptance rate (see Table \ref{tab: e2e_res}). We can conclude that the critical production data collected for GB use case was severely drifted.

\begin{table}[]
\caption{TWO-stage Filter Results}
\begin{tabular}{|c|c|l|l|l|l|}
\hline
Use case              & Encoding         & \multicolumn{1}{c|}{TA} & \multicolumn{1}{c|}{FA} & \multicolumn{1}{c|}{TR} & \multicolumn{1}{c|}{FR} \\ \hline
\multirow{3}{*}{POP} & VAE Enc          & 16.3\%                 & 2.0\%                  & 21.7\%                 & 60.0\%                 \\ \cline{2-6} 
                     & (VAE + Hist) Enc          & 50.3\%                 & 8.9\%                  & 14.9\%                 & 26.0\%                 \\ \cline{2-6} 
                     & Hist Enc & 21.4\%                 & 5.4\%                  & 18.3\%                 & 54.9\%                 \\ \hline
\multirow{3}{*}{GB}  & VAE Enc          & 31.2\%                 & 12.1\%                 & 20.7\%                 & 36.0\%                 \\ \cline{2-6} 
                     & (VAE + Hist) Enc         & 31.4\%                 & 5.5\%                  & 27.3\%                 & 35.9\%                 \\ \cline{2-6} 
                     &  Hist Enc & 11.9\%                 & 14.5\%                 & 18.2\%                 & 55.4\%                 \\ \hline
\end{tabular}
\label{tab: e2e_res}
\end{table}


\begin{tcolorbox} 
[colback=gray!8,colframe=gray!40!black]
\textbf{Finding:} The distribution-based filter successfully complements the confidence-based filter to lower false acceptances by up to six times, since it targets all accepted inputs with overconfident scores.
\end{tcolorbox}

\subsubsection*{RQ.4: Can CT on filtered data outperform vanilla CT in terms of F1-score enhancement after degradation?} 
\begin{figure}
   \centering
   \includegraphics[width=0.5\columnwidth]{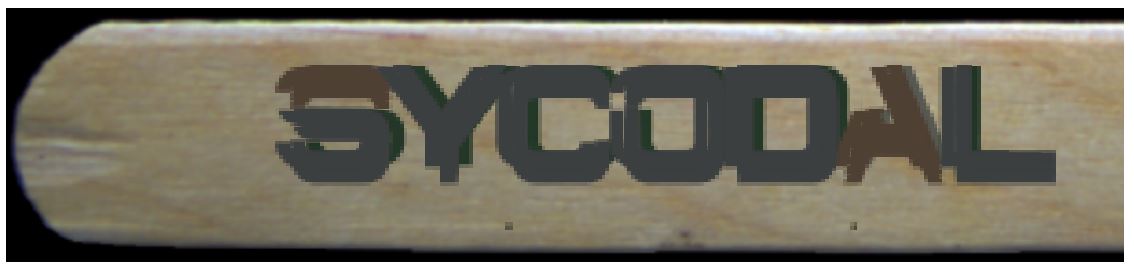}
   \caption{Smudged Stick Text Print}
   \label{fig:defaut_fonce}
   \vspace{-1em}
\end{figure}

\begin{figure}
\centering
\includegraphics[scale=0.06]{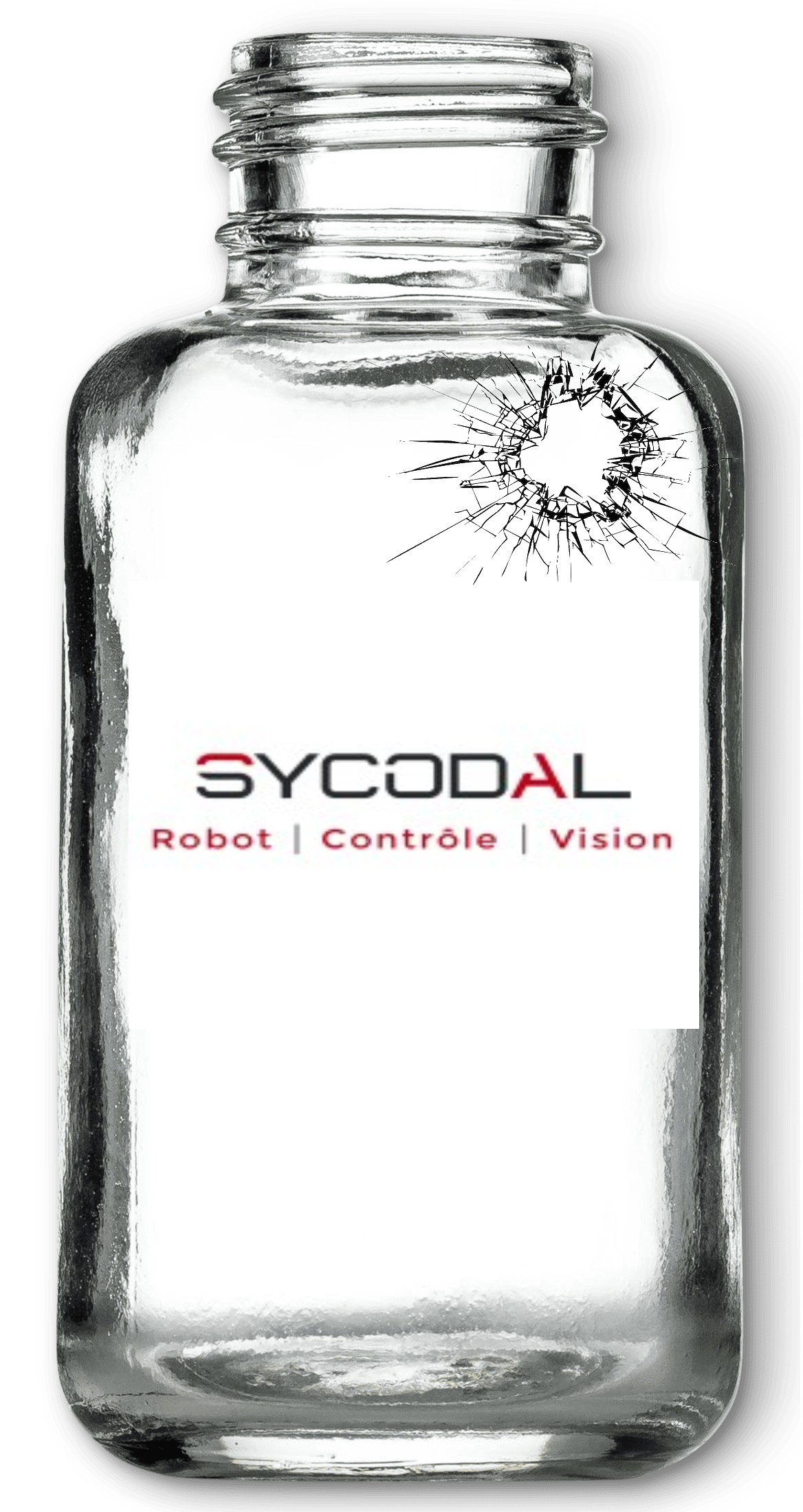}
\caption{Glass Bottle: Open Mark Defect}
\label{fig:gb_open_mark}
\vspace{-1em}
\end{figure}

\hfill
\\
\textbf{Motivation. }The main objective of the two-stage filter is to retain a valid subset of inputs that can be used to do CT-driven maintenance of the inspection system after performance degradation. Thus, we compare the performance improvement of the fine-tuned model with our proposed reliable CT to conventional (vanilla) CT. \\
\textbf{Method. }To perform CT, we propose to fine-tune the model in production using the novel data with self-generated labels, which can be divided into training (80\%) and validation (20\%) datasets. The difference between our proposed reliable CT and vanilla CT is the definition of novel data. A vanilla CT considers the production dataset collected within the last time window, i.e., since the last CT cycle. The reliable CT we propose uses a two-stage filtering strategy to remove risky instances from the production data, and feeds only a subset of the data into the model for fine-tuning. To compare the effectiveness of both CT approaches, we assess the performance of the fine-tuned in terms of F1-score based on two datasets: the first is the original validation set, which assesses the degree of induced catastrophic forgetting, and the second is the entire production set, which assesses the fine-tuned model's ability to cope with the latest conditions in the production environment that may remain for the foreseeable future. \\
\textbf{Results. } Table \ref{tab: ct_res} shows the results of validation F1 and production F1, which represent the F1 obtained by the fine-tuned model on the original validation dataset and the entire production dataset, respectively. The F1 scores are reported by CT approach, either vanilla or our proposal, and by industrial inspection use case. Table \ref{tab: ct_res} also includes the initial performance obtained by the original model as baseline for each use case. For POP use case, the original model has experienced a drastic decline in its F1 in production. According to CT results, vanilla approach slightly increased production F1 while almost decreasing validation F1 by the same percentage. Our proposed approach was able to increase production F1 more than it decreased validation F1. The analyses we made on our filtering results led us to conclude that the confidence scores have been diminished for POP, indicating that the model faces challenging inputs because most of the risky data has been discarded at the first filtering stage. Based on this, it is likely that the original model overfits the initial data and that its learned patterns won't hold up against the incoming inputs. Unfiltered self-generated labels may include a high number of false predictions, so Vanilla CT cannot fix the overfitting issue. In contrast, our reliable approach reduces the overfitting problem by selecting half of the production data with correct labels and high confidence to reinforce the model patterns at the deployment environment. In the GB case, vanilla CT not only failed to improve the fine-tuned model's performance but also caused a catastrophic forgetting with more than 13\% decay in the validation F1 data. In contrast, our proposed reliable CT succeeded to enhance both F1-score for original validation and production datasets. As we discussed during our analysis of filtering strategy assessment, the production data for GB shows severe drifts, leading to almost two-thirds of production inputs being rejected. Consequently, vanilla CT absorbs all production data associated with self-generated labels, leading to performance degradation on both original and novel distributions. As an alternative, our proposed approach relies only on a subset of data that remain close to the distribution and is confident that the self-generated labels will hold up, making the fine-tuned model more resilient against forgetting the original inductive bias and more capable of capturing the patterns required to cope with potential drifted inputs.\\
\textbf{Expert Feedback. } \\
\textit{POP Use Case Discussion: }Domain experts analyzed the images that the model could not handle. Based on the defects collected during the model development, pale-colored prints, decentralised texts, and erased logo parts are the most common issues. It is all related to incomplete or missing parts of the object of interest, which is the wood stick's logo or text. Model degradation occurred in production when the ink dispatcher of the printer machine was slightly damaged, resulting in smudged prints. This results in a new challenging type of defect (illustrated by Figure \ref{fig:defaut_fonce}) that is correlated with wrongly highlighted text or logo in a non appropriate way. This uncovered type of defect causes the model to behave poorly. Most of the images that were classified correctly have a mix of issues, as the prints were decentralised or there was uneven ink distribution between letters, which the original model can spot as defected. Despite correctly classified data being available, the volume was not sufficient to achieve acceptable performance through CT. Nevertheless, The fine-tuned model outperforms the original model on recent production data at the cost of performance degradation on the original validation data. Deploying the fine-tuned model is still advisable, as future data will likely match the recent distribution, ensuring the model adapts to current operational conditions. Meanwhile, alerting the maintenance team about the degradation allows them to investigate and address the root cause, maintaining overall model reliability.\\
\textit{GB Use Case Discussion: }Domain experts noticed two changes in the production environment. First, ambient lighting on-premises led to data shifts due to inadequate camera settings in terms of exposure. Second, a new defect was introduced in the production line, which resulted in a dramatic drift in input data. It is characterized by an open mark in the bottle's body, as shown in Figure \ref{fig:gb_open_mark}. The model predictions on these images are almost random due to the shifted location and form of defects, resulting in almost equal chances of correct and incorrect classifications. Thus, vanilla CT degrades both validation and production scores due to the prevalence of corrupted labels. The proposed approach, however, manages to discard the majority of these new defected products and fine-tune the model based on images that are shifted in brightness, resulting in an increased model robustness. In this use case, it outperforms the vanilla CT with a modest improvement. Importantly, by trimming the risky data, we prevent the catastrophic forgetting that causes substantial degradation in the fine-tuned model's performance on the original validation data when using vanilla CT. This careful data filtering strategy ensures more stable and reliable performance over time.\\

\begin{tcolorbox} 
[colback=gray!8,colframe=gray!40!black]
\textbf{Finding:} Our proposed CT-driven maintenance approach strengthens the positive impact of vanilla CT, while reducing the risk of performance degradation caused by high false prediction rates.
\end{tcolorbox}

\begin{table}[]
\centering
\caption{Continuous Training Results}
\begin{tabular}{|l|l|l|l|}
\hline
Use case              &               & Validation F1 & Production F1 \\ \hline
\multirow{3}{*}{POP} & Original Model & 93.2\%        & 45.8\%        \\ \cline{2-4} 
                     & Vanilla CT     & 91.7\%       & 48.8\%        \\ \cline{2-4} 
                     & Our Approach   & 88.0\%          & 59.7\%        \\ \hline
\multirow{3}{*}{GB}  & Original Model & 93.3\%        & 76.6\%       \\ \cline{2-4} 
                     & Vanilla CT     & 80.2\%        & 71.4\%        \\ \cline{2-4} 
                     & Our Approach   & 93.8\%        & 79.1\%        \\ \hline
\end{tabular}
\label{tab: ct_res}
\end{table}

\section{Related work}
\label{sec:related_work}
Over the last few years, with the wide adoption of ML models in dynamic environments, extensive research has been conducted to ensure model quality after deployment.\\
For data validity monitoring, drift detection approaches are proposed by employing either model certainty or data distribution as indicators. Confidence metrics are often leveraged, operating on the assumption that higher confidence implies more accurate predictions, indicating alignment with the known environment setting. For instance, \cite{feutry2019simple} compares confidence scores, i.e., softmax logits, to pre-defined thresholds, and \cite{alberge2019detecting} conducts tests that compare original confidence distributions (i.e., softmax logits) with new distribution logits. Similarly, \cite{ackerman2021automatically} perform two tests on two distributions: confidence of the predicted class and confidence for all other classes. To use calibrated confidences, \cite{ginsberg2022learning} opts for Constraint Disagreement Classifiers (CDCs) to estimate the aggregated cross-entropy, i.e., confidence, and \cite{lee2017training, bourboux2022trustgan} leverage confidence calibration by data augmenting using generative models. Our approach leverages confidence scores to create a confidence-based filter, but we make this the first filtering step that eliminates self-aware unreliable predictions. Even so, we believe these confidence scores will suffer from degeneration in reliability if the input data substantially differ from the initial samples used in calibrating and assessing the uncertainty/confidence.\\
To identify potential drifts in imaging data, dimensionality reduction is necessary since distributional analysis is more effective with a low-dimensional representation. Generative models such as AEs and VAEs \cite{rabanser2019failing, soin2022chexstray} can be used to derive latent features, thereby encoding the most relevant factors for reconstructing images.
In addition to exploring geometric distances within the dimensionally-reduced space, alternative strategies opt for statistical tests such as Kolmogorov-Smirnov (KS), Chi-square, or Maximum Mean Discrepancy (MMD) \cite{rabanser2019failing}, as well as two-sample testing \cite{soin2022chexstray}. \cite{nikolov2021seasons} performs statistical testing on the reconstruction loss from VAE aggregated with domain variable. Nevertheless, statistical tests often assume an underlying data distribution and can be sensitive to sample size and number of dimensions. \cite{ugrenovic2021towards} also combines loss with domain variables, but feeds them into an OCC model to capture the pertinent in-distribution characteristics, allowing the systematic elimination of drifted inputs. Our research suggests that loss alone may not adequately capture shifts, so we advocate the use of the full embedding to capture shifts. We propose a Histogram encoding technique to complement the model-generated embeddings. Indeed, the Histogram provides fine-grained pixel information that can assist in capturing shifts in image pixel characteristics, not semantic-level changes.\\
Continual Learning (CL) under data drifts was studied from the algorithmic perspective such as enabling the adaptive learning of emerging new classes \cite{zhang2023toward}. The purpose of our approach is to improve the continuous training of DL models using a data-driven approach based on advances in uncertainty quantification and drift detection to select more reliably the input data samples for CT. In particular, our approach effectively eliminates non-representative, unusual ID entries, substantially shifted entries, and unexpected, anomalous entries from the fresh data collected in production settings.

\section{Threats to Validity}
\label{sec:threats_to_validity}
In this paper, we proposed an approach to enhance the automated selection of relevant production data for CT purposes. Our approach demonstrated its efficiency across two use cases in critical scenarios. Despite the promising results, several threats to validity exist.  \\
\textbf{External Threats.} Our experiments are limited to  visual inspection, which limits the generalizability of results. To address this, we opted for two different use cases POP and GB that are different in terms of task complexity, product, texture, dimension and defect types and we repeated each execution at least 5 times. \\
\textbf{Internal threats.} Confidence filter design involves the use of softmax despite the existence of multiple confidence scores such as entropy, perplexity score etc. This choice is being motivated by the wide use and proven performance of softmax in classification tasks \cite{pearce2021understanding}, and to address its limits regarding overconfidence, we opted for calibration. Since the absence of drifted data by design and to enhance models generalization we opted for image transformation augmentation known for being unrealistic \cite{gontijo2020affinity}, to address this we relied on domain experts to define transformations and respective parameters. Finally, although our approach requires domain-expert involvement during approach setup, this involvement is not frequently needed and it ensures setup reliability.

\section{Conclusion}
\label{sec:conclusion}
We introduce a novel two-stage filter to trim the risks in CT using self-generated labels in dynamic production environment. First, we use calibrated confidence scores to eliminate risky data with high uncertain predictions. Second, we employ VAE and Histogram to encode rich image embeddings serving for OOD model optimization to filter significantly shifted risky data. The accepted subset is suitable for reliable  maintenance of inspection system models. Our evaluation on two industrial inspection systems demonstrates that our approach passes less than 9\% of false predictions, leading to an increase of F1-score up to 14\% on critical production data. We discover limitations of supervised learning algorithms such as their inability to self-learn new defects and their reliance on human verification. Even if our reliable CT approach is applied for successive cycles, the model could not capture new defect patterns that are not similar to the ones from in-distribution. In the future work, we will study the unsupervised anomaly detection models for industrial inspection and their automated maintenance and evolution in terms of adaptability to dynamic environments and natural recognition of novel defects.

\bibliographystyle{plainnat}
\bibliography{refs}

\end{document}